%% file: arxiv.tex
\documentclass[letterpaper]{article} 
\usepackage[preprint]{aaai2027-arxiv}  
\usepackage[hyphens]{url}  
\usepackage{graphicx} 
\usepackage{natbib}  
\usepackage{caption} 
\usepackage{algorithm}
\usepackage{algorithmic}

\usepackage{newfloat}
\usepackage{listings}
\DeclareCaptionStyle{ruled}{labelfont=normalfont,labelsep=colon,strut=off} 
\floatstyle{ruled}
\newfloat{listing}{tb}{lst}{}
\floatname{listing}{Listing}

\input{math_commands.tex}

\usepackage{booktabs}
\usepackage{graphicx}        
\usepackage{multirow}        
\usepackage{xcolor}          
\usepackage{microtype}       
\usepackage{xspace}          

\usepackage{mathtools}

\newcommand{\Lpretrain}{\mathcal{L}_{\mathrm{pre}}}
\newcommand{\Ljoint}{\mathcal{L}_{\mathrm{joint}}}
\newcommand{\Lvid}{\mathcal{L}_{\mathrm{vid}}}

\newcommand{\dvrk}{dVRK\xspace}
\newcommand{\surrol}{SurRoL\xspace}
\newcommand{\cosmospol}{Cosmos Policy\xspace}

\usepackage[colorlinks=true,linkcolor=blue,citecolor=blue,urlcolor=blue,breaklinks=true]{hyperref}
\hypersetup{
  pdftitle={Surgical WAM: A World-Action Model for Data-Efficient Surgical Robot Learning},
  pdfauthor={Wenrui Bao, Tianyun Jiang, Zhiben Chen, Ser-Nam Lim, Peter D. Peng, Yuzhang Shang}
}
\usepackage{cleveref}        
\crefname{section}{Section}{Sections}
\crefname{figure}{Figure}{Figures}
\crefname{table}{Table}{Tables}
\crefname{equation}{Equation}{Equations}
\crefname{enumi}{item}{items}
\usepackage[export]{adjustbox}
\usepackage{bm}

\title{Surgical WAM: A World-Action Model for Data-Efficient Surgical Robot Learning}

\newcommand{\surgwam}{\textbf{Surgical WAM}\xspace}

\author{
    Wenrui Bao\textsuperscript{\rm 1},
    Tianyun Jiang,
    Zhiben Chen,
    Ser-Nam Lim\textsuperscript{\rm 1},
    Peter D. Peng\textsuperscript{\rm 2},
    Yuzhang Shang\textsuperscript{\rm 1}\corresponding
}
\affiliations{
    \textsuperscript{\rm 1}University of Central Florida,    \textsuperscript{\rm 2}AdventHealth
}

\begin{document}

\maketitle

\input{sec/abs}

\input{sec/intro}
\input{sec/related}
\input{sec/method}
\input{sec/exp}

\input{sec/con}




\bibliography{aaai2027}
\newpage


\end{document}

%% file: math_commands.tex
\usepackage{amsmath,amsfonts,bm}

\newcommand{\figtop}{{\em (Top)}}
\newcommand{\figbottom}{{\em (Bottom)}}

\def\eqref#1{equation~\ref{#1}}

\def\1{\bm{1}}

\DeclareMathAlphabet{\mathsfit}{\encodingdefault}{\sfdefault}{m}{sl}
\SetMathAlphabet{\mathsfit}{bold}{\encodingdefault}{\sfdefault}{bx}{n}



%% file: sec/abs.tex
\begin{abstract}
Learning reliable surgical manipulation policies is bottlenecked by the scarcity of action-labeled demonstrations: teleoperated surgical robot (e.g., \dvrk{}) trajectories with synchronized kinematics are costly to collect, while surgical tasks demand precise contact handling, long-horizon reasoning, and bimanual coordination. Endoscopic video is comparatively inexpensive and abundant relative to synchronized video--kinematics trajectories, and a natural way to exploit it is to learn world models of surgical scenes. However, existing surgical world models use video primarily for simulation or policy evaluation, and rarely translate the learned dynamics into closed-loop control. This gap raises our central question: under a fixed budget of action-labeled demonstrations, does action-free video pretraining improve closed-loop surgical manipulation? To answer it, we introduce the Surgical World-Action Model (\surgwam), a unified generative model built on Cosmos Policy that jointly predicts future endoscopic observations and executable surgical robot action chunks. \surgwam first learns surgical visual dynamics from action-free video and is then fine-tuned on the fixed action-labeled budget; at deployment, it acts as a closed-loop, receding-horizon controller that executes a short prefix of each predicted action chunk and replans from the resulting observation. On a suite of four simulated surgical manipulation tasks, video pretraining improves the average success rate from $63.5\%$ to $77.8\%$, including an absolute gain of $20$ percentage points on PegTransfer, with the largest improvements on contact-rich and bimanual tasks. These results demonstrate that action-free video provides transferable visual dynamics priors for learning surgical robot control with limited action supervision, positioning data-efficient video pretraining as a practical path toward scaling up surgical robot learning.


\end{abstract}

%% file: sec/intro.tex
\section{Introduction}
\label{sec:intro}

Autonomous surgical robots could improve the consistency, accessibility, and scalability of minimally invasive surgery~\citep{yang2017medical,dupont2021decade,attanasio2021autonomy}. However, learning reliable surgical manipulation policies remains challenging. A successful policy must achieve sub-millimeter precision, reason about contact with deformable tissue, and perform long-horizon actions whose success is often observable only at the end of a procedure, such as suturing or needle picking~\citep{kassahun2016surgical,ostrander2024suturing}.

A central obstacle is the scarcity of action-labeled surgical robot data. Collecting teleoperated \dvrk{} trajectories with synchronized kinematic labels requires access to surgical hardware, specialized environments, expert operators, and task-specific setup. Consequently, existing surgical manipulation datasets typically contain only hundreds to a few thousand demonstrations~\citep{haworth2025suturebot,kim2025srth}, far fewer than the millions of frames commonly used in general-purpose robot learning~\citep{walke2023bridgedata,oneill2024openx,khazatsky2024droid}. This scarcity limits both the diversity of behaviors that can be learned and the robustness of the resulting policies.

World models provide a potential way to alleviate this limitation. By predicting future observations conditioned on actions, they learn temporal dynamics and task-relevant physical structure~\citep{ha2018worldmodels,hafner2020dreamer}. Unlike purely reactive Vision-Language-Action (VLA) policies~\citep{brohan2023rt2,kim2024openvla,black2024pi0}, predictive models can provide forward-looking representations of the consequences of candidate actions~\citep{shen2025videovla,pai2025mimicvideo}. Importantly, the observation-prediction objective can be trained on video without action labels, allowing representation learning to use data that would otherwise be unsuitable for policy training.

\begin{figure*}[!t]
\centering
\includegraphics[width=1\textwidth]{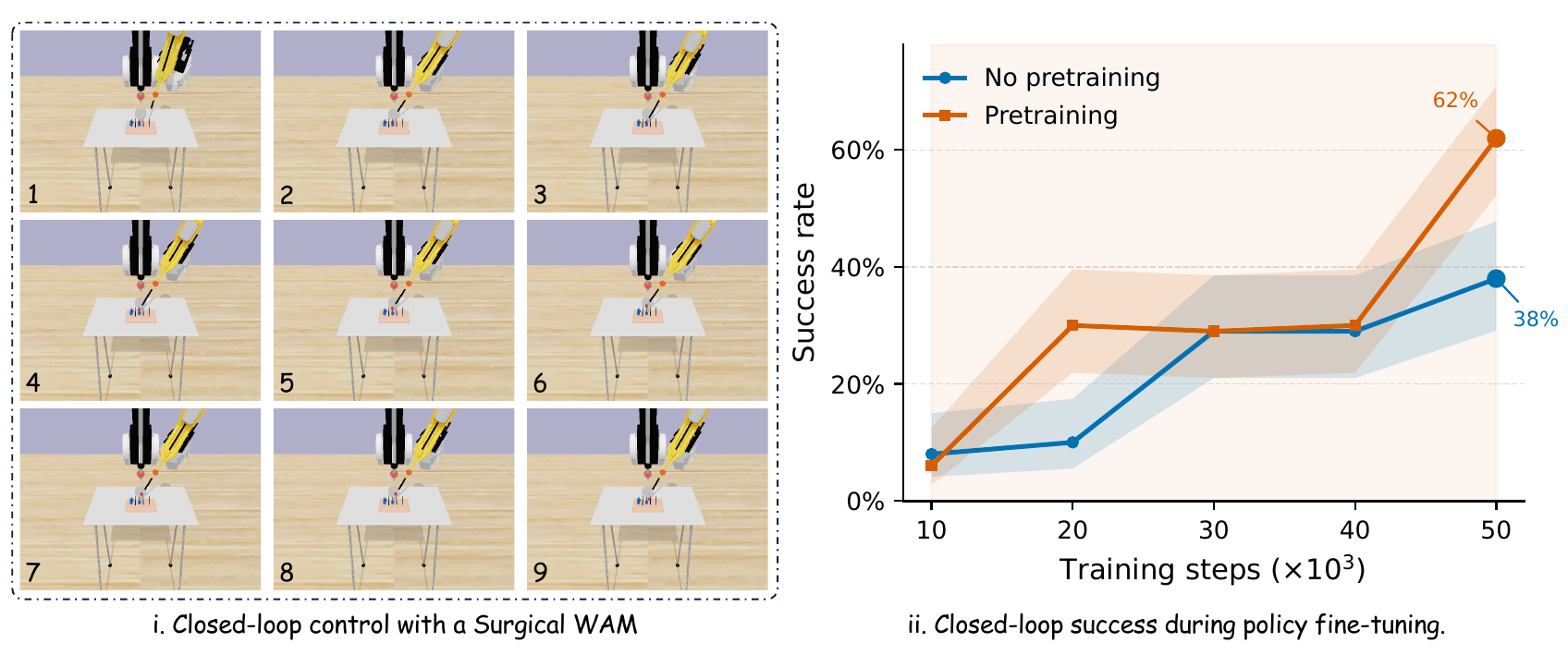}
\vspace{-3mm}
\caption{\textbf{\surgwam drives closed-loop \dvrk{} manipulation and makes policy learning markedly more efficient.}
\textbf{(i)} A \dvrk{} robot driven by \surgwam performs the PegTransfer task in closed loop. At each control step, the model receives the current endoscopic observation and \dvrk{} proprioception, jointly predicts future observations and an action chunk, executes a short prefix of the predicted actions, and replans from the resulting observation, progressing through reaching, grasping, lifting, transferring, and placing.
\textbf{(ii)} The proposed action-free video pretraining substantially improves training efficiency: under the same action-labeled budget, the pretrained model attains a higher peak success rate with fewer fine-tuning steps (62\% at 50k steps vs. 50\% at 60k steps without pretraining, averaged over the \surrol{} task suite).}
\vspace{-4mm}
\label{fig:teaser}
\end{figure*}

Recent World-Action Models (WAMs) integrate these two capabilities into a single model. A WAM jointly predicts future video and the robot actions that would produce the predicted future, thereby sharing representations between world modeling and control~\citep{ye2026dreamzero}. This joint formulation enables the two objectives to complement each other: generated rollouts inform action prediction, while action supervision focuses video prediction on task-relevant objects and dynamics. It also avoids the need to separately train and coordinate a world model and a control policy. For example, \cosmospol{}~\citep{kim2026cosmospolicy} adapts a latent video diffusion model by adding an action-prediction output and achieves competitive manipulation performance without introducing a separate policy architecture.

Despite this progress, world models in surgical robotics have primarily been used in two separate roles. First, they have been used as evaluators that generate predicted video for assessing externally trained policies, without directly controlling the robot~\citep{zbinden2025cosmossurgdvrk}. Second, they have been used as synthetic-data generators in multi-stage pipelines, where a world model generates surgical videos, an inverse-dynamics model assigns pseudo-kinematics, and a separate VLA policy is trained on the resulting data~\citep{he2026cosmoshsurgical}. These approaches do not exploit the shared representation between prediction and control that is central to the WAM formulation. Moreover, they are commonly evaluated using open-loop prediction or trajectory metrics rather than closed-loop task success.

This gap motivates our central question: \textit{under a fixed budget of action-labeled demonstrations, does action-free video pretraining improve closed-loop surgical manipulation?} The question is particularly relevant to surgery because video and action labels have substantially different collection costs. Endoscopic video can be collected from procedural recordings or generated in simulation, whereas paired robot kinematics require specialized hardware, instrumentation, and expert teleoperation. A WAM provides a direct mechanism for exploiting this asymmetry: its world-model component can be pretrained using video alone, while the action head can later be adapted using a comparatively small action-labeled dataset. If video pretraining improves the learned representation of surgical dynamics, it should improve downstream closed-loop control without increasing the action-label budget.

In this work, we propose \surgwam, a closed-loop world-action model for surgical manipulation. At each control step, the model takes the current endoscopic observation and \dvrk{} proprioceptive state, and jointly predicts a future video sequence together with an action chunk; it executes a short prefix of the chunk and replans from the resulting observation. This receding-horizon loop lets the model continually correct its predictions and actions as the scene evolves (\Cref{fig:teaser}(i)). We evaluate \surgwam{} through closed-loop execution on four \dvrk{} manipulation tasks in the \surrol{} benchmark~\citep{xu2021surrol}, which together cover unimanual, bimanual, and contact-rich behaviors such as reaching, grasping, and coordinated object transfer.


To test whether \surgwam{} can acquire useful surgical dynamics from action-free video, we hold the action-labeled budget fixed and vary only the volume of video-only pretraining data. Under this controlled setting, action-free video pretraining consistently improves closed-loop task success and substantially accelerates fine-tuning: as shown in \Cref{fig:teaser}(ii), it reaches a higher peak success rate with fewer fine-tuning steps than training on action labels alone. Benchmark and task details are given in \Cref{sec:experiments}.


Overall, this work makes three contributions.

First, we develop \surgwam, to our knowledge the first world-action model applied to surgical robot learning. The model jointly predicts future endoscopic video and \dvrk{} actions within a single architecture and directly drives closed-loop task execution.

Second, we answer the paper's central question: under a fixed budget of action-labeled demonstrations, action-free video pretraining improves closed-loop surgical manipulation. Fine-tuned on the same action-labeled data with the same architecture, the video-pretrained WAM achieves higher closed-loop success and reaches its peak with substantially fewer fine-tuning steps; the gain thus comes entirely from the video pretraining, at no extra cost in action supervision.

Third, we characterize when this benefit is realized: ablations show that the video-pretrained model peaks with half the fine-tuning steps, is robust across execution horizons, and gains most on contact-rich and bimanual tasks. The same trend holds on real \dvrk{} video from JIGSAWS, confirming the learned prior captures genuine surgical dynamics rather than a simulator artifact.

Together, these findings position data-efficient, action-free video pretraining as a practical path toward scaling up surgical robot learning: the bottleneck shifts from collecting expensive teleoperated demonstrations to curating video that the surgical domain already produces at scale.

%% file: sec/related.tex
\section{Related Work}
\label{sec:related}

\subsection{Surgical Robot Learning and Benchmarks}
\label{sec:related_surgical}
Surgical robot learning has progressed from model-based autonomy toward data-driven control. Early systems such as STAR combined explicit perception, tracking, planning, and control to demonstrate supervised-autonomous suturing and laparoscopic anastomosis~\citep{shademan2016star,saeidi2022star}. The open \dvrk platform~\citep{kazanzides2014dvrk} and datasets such as JIGSAWS~\citep{gao2014jigsaws} subsequently made synchronized surgical video and kinematic data more accessible for learning-based methods. In simulation, \surrol~\citep{xu2021surrol} has become a widely used \dvrk-compatible benchmark for goal-conditioned surgical manipulation, alongside newer environments designed for higher-throughput data collection and deformable-tissue interaction~\citep{yu2024orbitsurgical,scheikl2023lapgym,schmidgall2024surgicalgym}. These platforms make surgical robotics a useful testbed for learning-based control.

Policies trained in these settings commonly use goal-conditioned reinforcement learning, demonstration-augmented reinforcement learning, or imitation learning~\citep{andrychowicz2017her,nair2018demonstrations,huang2023dex,goecks2020col}. Recent systems have also extended data-driven learning to real surgical robots through hierarchical imitation learning and language-conditioned vision-language-action policies \citep{haworth2025suturebot,kim2025srth}. These methods establish important benchmarks, datasets, and policy baselines, but they learn directly from state-action or image-action pairs and thus cannot exploit surgical video that lacks kinematic labels. Our work instead treats such action-free video as a first-class training resource: we use it to pretrain a generative model of surgical dynamics and show that this improves closed-loop control on the same \surrol{} tasks under a fixed action-label budget.

\subsection{Video-Conditioned Control and World-Action Models}
\label{sec:related_wam}
Video-conditioned robot learning has emerged as a way to reduce the dependence on action-labeled data. Earlier approaches typically separated visual planning from control: a model first generated a visual subgoal or trajectory, after which a downstream policy or inverse model converted the prediction into robot actions~\citep{du2023unipi,ko2024avdc, black2024susie}. Other approaches use video prediction to learn predictive visual representations or to initialize general-purpose robot models~\citep{wu2024gr1,hu2025vpp, tian2025seer}. Interactive video models further extend this idea by modeling how observations evolve under robot interaction \citep{wu2024ivideogpt,bruce2024genie,agarwal2025cosmos}. The shared motivation is that visual prediction encourages models to represent object motion, contact events, and task progress from video that does not necessarily contain action labels.

A limitation of many video-conditioned systems is that video and action prediction remain separate stages. A visual model may generate a plausible future without representing which actions would produce it, while an inverse model may convert a predicted video into actions without sharing the representation used for visual prediction. This separation can be especially problematic for surgical manipulation, where small geometric errors may change contact outcomes. Recent World--Action Models (WAMs) instead unify future-state prediction and action generation in a single generative model \citep{zhu2025uwm,shen2025videovla,pai2025mimicvideo,yuan2026fastwam}. By sharing representations between the two streams, WAMs allow visual dynamics learned from video to inform control while using action supervision to align prediction with executable behavior.

To our knowledge, ours is the first world--action model applied to surgical manipulation. Rather than proposing a new general-purpose architecture, we instantiate the paradigm with Cosmos Policy~\citep{kim2026cosmospolicy}, which generates future visual content and robot actions in a shared latent sequence on top of a latent video diffusion model. This brings the WAMs to the domain where action labels are among the scarcest and control requirements are most stringent, and we quantify its benefit through closed-loop execution.

\begin{figure*}[t]
\centering
\includegraphics[width=0.99\textwidth]{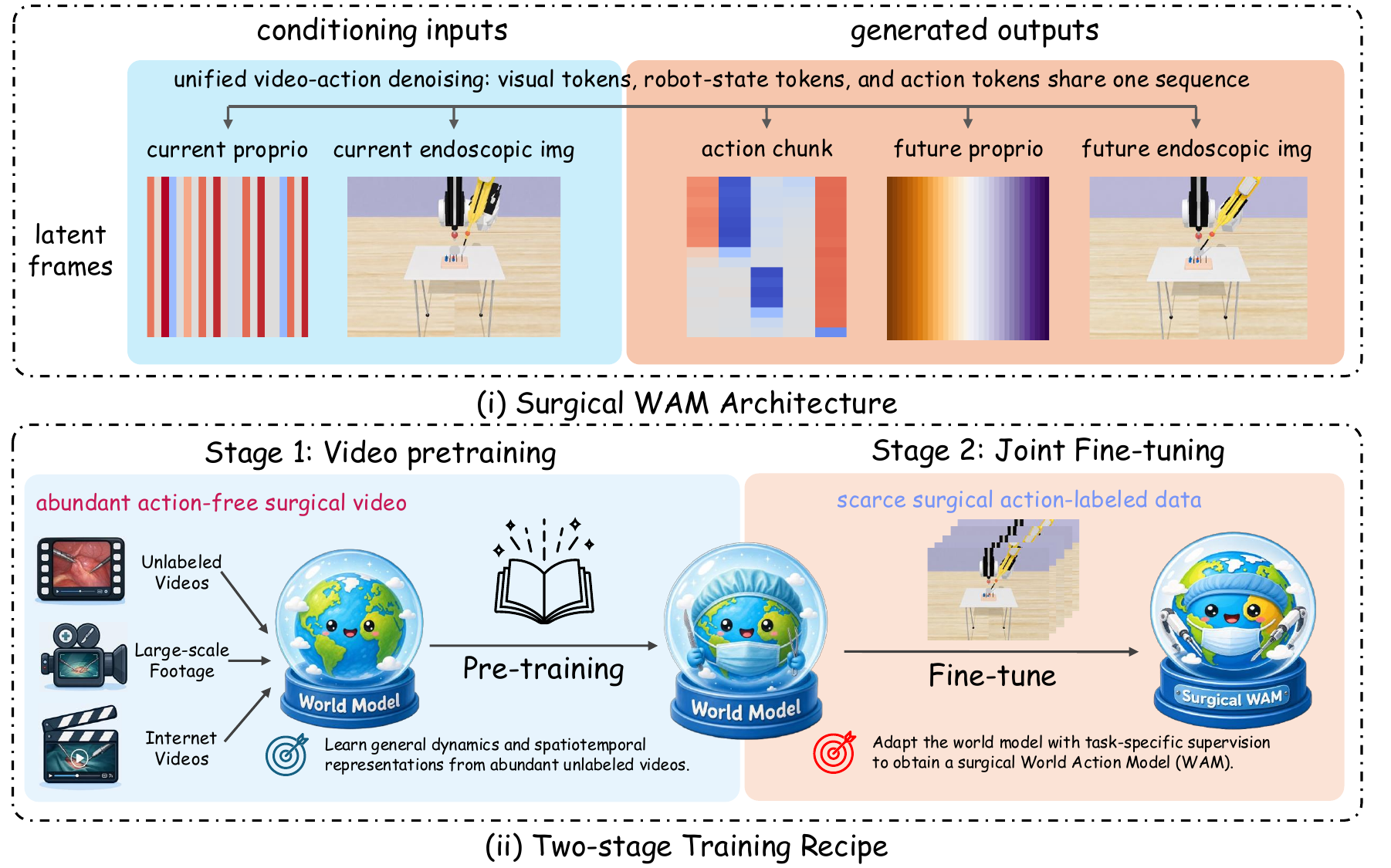}
\vspace{-1mm}
\caption{\textbf{Surgical WAM architecture and two-stage training recipe.}
\figtop{} \textit{Architecture.} Closed-loop inference with the \surgwam on SurRoL peg transfer. At each control step, the model conditions on the current endoscopic frame, robot state and goal, and the task text. A single video-action diffusion transformer samples an action chunk in the same latent sequence used for visual/state prediction. \figbottom{} \textit{Two-stage training recipe.} \textbf{Stage 1 (Video pretraining):} the world model is fine-tuned on unlabeled surgical video using only the video loss $\Lvid$; action latent frames are masked (zeroed). \textbf{Stage 2 (Joint fine-tuning):} the pretrained world model is jointly fine-tuned on action-labeled demonstrations using the full objective $\Ljoint$.}
\vspace{-4mm}
\label{fig:arch}
\end{figure*}

\subsection{World and Video Models for Surgery}
\label{sec:related_surgworld}
Generative modeling has recently been applied to surgical video synthesis, geometric reconstruction, and surgical foundation models. Endoscopic video generators condition on text, motion, or procedural structure to render plausible surgical scenes and future observations \citep{li2024endora,chen2025surgsora}. Neural rendering and digital-twin methods reconstruct surgical scenes, including deformable tissue and endoscopic viewpoints~\citep{wang2022endonerf,liu2024endogaussian}. Other surgical foundation models focus on visual question answering, procedural reasoning, and vision-language assistance~\citep{zeng2025surgvlm}. These approaches demonstrate the value of domain-specific generative representations, but their evaluations primarily focus on visual synthesis, perception, reasoning, or open-loop prediction.

The closest surgical applications of Cosmos-family models also use interfaces that differ from ours. Cosmos-Surg-dVRK predicts future frames under candidate actions to evaluate externally trained policies, rather than generating actions for the robot~\citep{zbinden2025cosmossurgdvrk}. Cosmos-H-Surgical/ SurgWorld uses a video-only world model followed by inverse-dynamics modeling to produce pseudo-labeled data for a separate open-loop VLA policy \citep{he2026cosmoshsurgical}. SAW similarly focuses on action-conditioned surgical video synthesis without integrating an action head into the same generative model or evaluating closed-loop task execution \citep{rapuri2026sawsurgicalactionworld}.

In contrast, our \surgwam keeps prediction and control in a single shared generative model: it jointly predicts future endoscopic observations and executable \dvrk{} actions, without relying on an external policy, inverse-dynamics model, or synthetic-data pipeline. We evaluate it directly through receding-horizon closed-loop manipulation on \surrol{} and isolate the contribution of action-free video pretraining under a fixed action-label budget. In short, prior surgical world models ask whether plausible surgical video can be generated or scored; we test whether the same visual-dynamics prior can be made to act.

%% file: sec/method.tex
\section{Method}
\label{sec:method}

Our goal is to learn a closed-loop surgical manipulation policy under a
fixed budget of action-labeled demonstrations, while additionally
exploiting surgical video that carries no action labels. We realize this
with a two-stage recipe built on the world--action model paradigm
(\Cref{fig:arch}): the model first acquires surgical visual dynamics
through action-free video pretraining, and is then fine-tuned jointly on
video and action prediction using the fixed action-labeled budget. This
section introduces the model class and notation (\Cref{sec:prelim}), the
two training stages (\Cref{sec:recipe}), and closed-loop execution at
test time (\Cref{sec:inference}).

\subsection{Surgical World-Action Model}
\label{sec:prelim}

\paragraph{World--action models.}
A recent line of work trains a single network to jointly model future
observations and future actions, rather than separating world modeling
from policy learning~\citep{shen2025videovla,pai2025mimicvideo,
yuan2026fastwam,kim2026cosmospolicy}. \citet{ye2026dreamzero} group
these models under the term \emph{world--action model} (WAM), which we
adopt as a descriptive label for the joint distribution
\begin{equation}
  p_\theta\!\left(o_{t+1:t+K},\, a_{t:t+H_c}\,\middle|\,o_{\le t},\, s_{\le t},\, c\right),
  \label{eq:wam}
\end{equation}
where $o_{\le t}$ and $s_{\le t}$ are the visual and proprioceptive
history, $c$ is a task embedding, $o_{t+1:t+K}$ are $K$ future
predictions, and $a_{t:t+H_c}$ is an action
chunk~\citep{zhao2023act,chi2023diffusionpolicy} of length $H_c$. In our
surgical setting, $o_t$ is the current endoscopic RGB frame and $s_t$
the \dvrk{} proprioceptive state.
Our recipe applies to any WAM whose latent sequence
contains both video and action slots processed by a shared network,
so that weights learned without action labels can be reused for
action prediction.

\paragraph{Instantiation.}
As the backbone used throughout our
experiments, we instantiate the WAM with Cosmos
Policy~\citep{kim2026cosmospolicy}, a diffusion-transformer
realization of \cref{eq:wam} whose latent sequence contains dedicated
slots for past visual observations, robot state, future predictions,
and actions, all processed by a single network. Its visual and
future-prediction slots are architecturally compatible with
action-free video diffusion backbones, so weights obtained by
action-free video pretraining transfer directly into Cosmos Policy
and serve as the initialization for the action-labeled stage.

\subsection{Two-Stage Training}
\label{sec:recipe}

Our training procedure is informed by recent evidence on the role of
the video-prediction objective in a WAM.
\citet{yuan2026fastwam} show that removing video co-training
substantially degrades manipulation performance, and predictive
visual representations have likewise been shown to transfer to
control~\citep{hu2025vpp,shen2025videovla,pai2025mimicvideo}.
Crucially, acquiring such a representation does not require
action-labeled data: it can be learned from action-free video alone,
reserving the scarce action labels for grounding the learned
dynamics in executable commands. We therefore decouple training into
a video-only pretraining stage followed by action-labeled
fine-tuning.

\paragraph{Stage 1: action-free video pretraining.}
The first stage trains only the world-model component. Given the visual
history of an action-free clip, the model learns to denoise the
future-video latents, while the action and state slots are excluded
from both input and supervision:
\begin{equation}
\Lpretrain(\theta)
=
\mathbb{E}_{\,o\sim\mathcal{D}_{v},\,\epsilon,\,\tau}
\left[
\left\|
\epsilon
-
\epsilon_{\theta}\!\left(z^{\mathrm{vid}}_{\tau},\tau \mid o_{\le t}\right)
\right\|_2^2
\right],
\label{eq:lpre}
\end{equation}
where $\mathcal{D}_{v}$ is an action-free video corpus and
$z^{\mathrm{vid}}_{\tau}$ denotes the noised future-video latents at
diffusion timestep $\tau$. Because \cref{eq:lpre} requires no kinematic
labels, it can consume surgical video at a scale unreachable by
teleoperated data collection. In this work we instantiate Stage 1 with
the surgical video world model of \citet{he2026cosmoshsurgical},
obtained by fine-tuning a general video diffusion
backbone~\citep{agarwal2025cosmos} on large-scale surgical video. Its weights populate the visual and
future-prediction slots of the WAM, while the action and state slots
are initialized fresh and learned in Stage 2.

\paragraph{Why the recipe fits surgery.}

Inside a WAM, learning splits into two problems with very different
data costs. Learning how the endoscopic scene evolves requires only
video, which surgery produces abundantly, whereas learning which
command realizes a desired future requires synchronized kinematics,
which are scarce and expensive. Training both from labeled
demonstrations alone ignores this asymmetry, because the costly
labels must then also teach visual dynamics that unlabeled video
could have taught instead. The two-stage recipe assigns each problem
to the data source that suits it: abundant video builds the dynamics
representation in Stage 1, and the scarce labels are spent solely on
grounding action prediction in that representation in Stage 2.

\begin{table*}[t]
\caption{\textbf{Main results on the \surrol{} benchmark.} We report
closed-loop task success rates (\%) on four surgical manipulation
tasks (snapshots shown in the top row). All methods are trained
on the same action-labeled dataset with an identical fine-tuning
budget. Best results
are in \textbf{bold}. BET: Behavior Transformer; DEX: demonstration-guided
exploration; PT: video pre-training.}
\vspace{-0.1in}
\label{tab:main}
\centering
\small
\begin{tabular}{l|cccc}
\toprule
\textbf{Model} & Needle Pick & Peg Transfer & Needle Regrasp & BiPeg Transfer \\
\midrule
snapshots & \includegraphics[width=0.15\linewidth, valign=c]{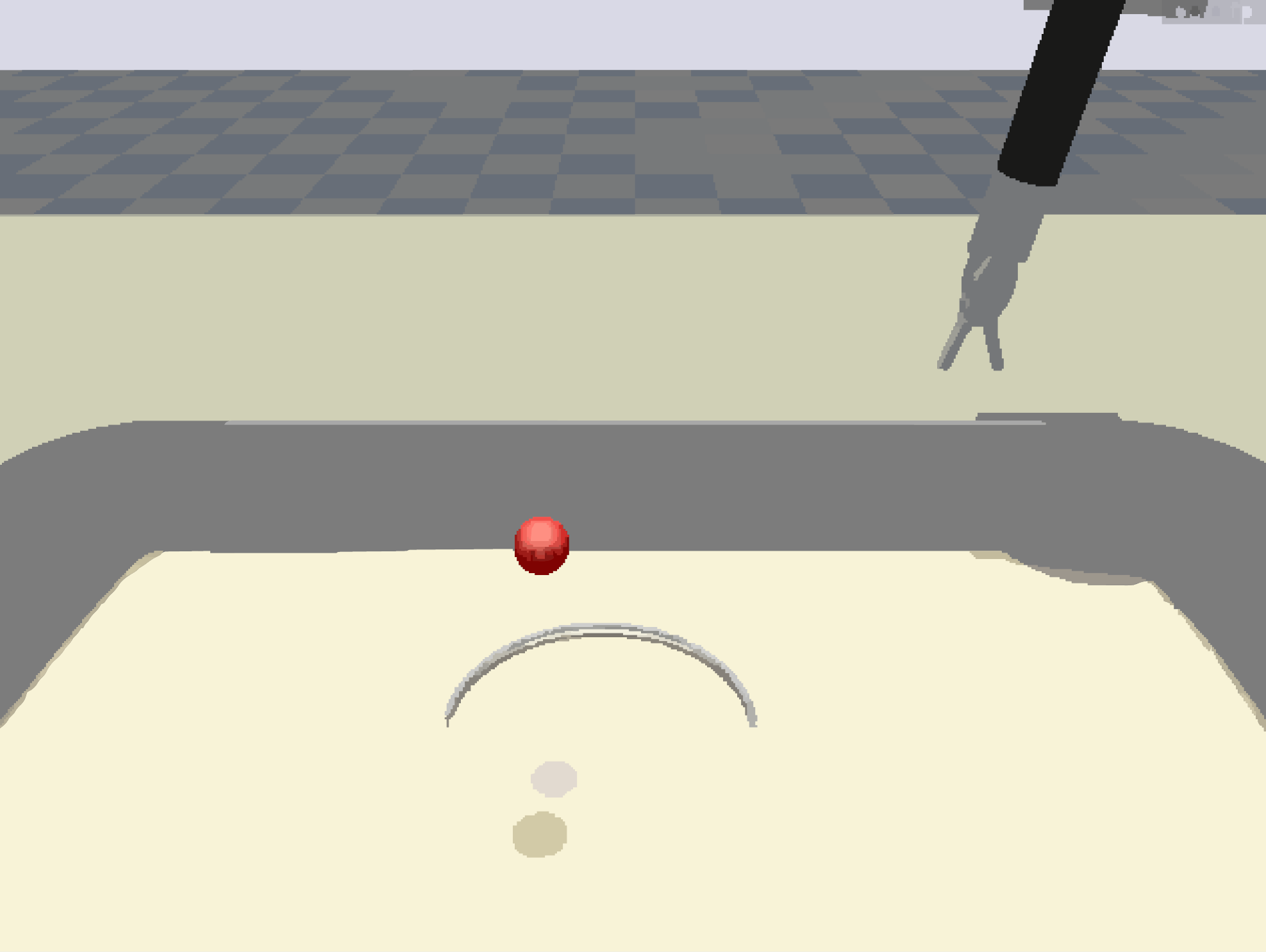} & \includegraphics[width=0.15\linewidth, valign=c]{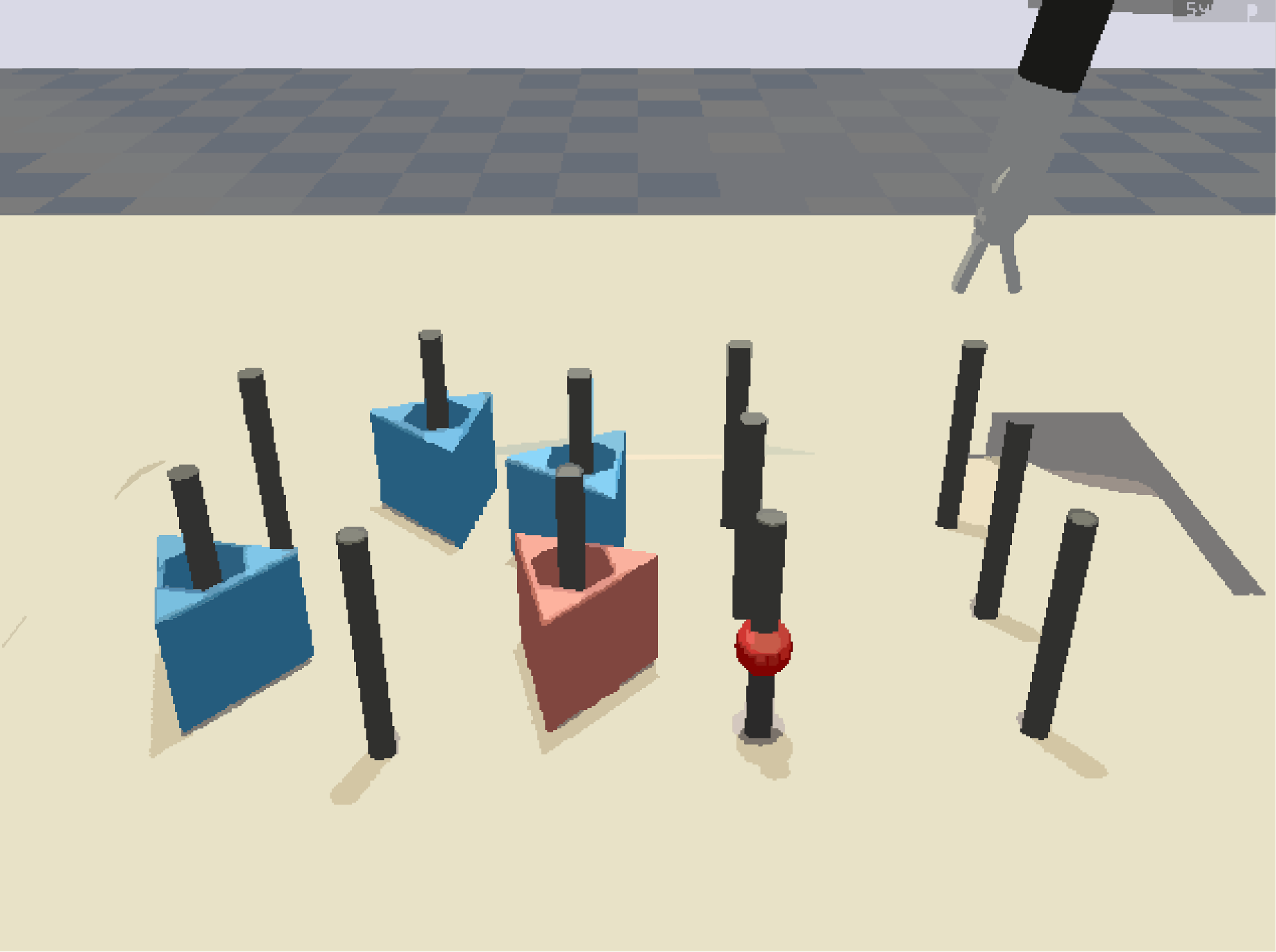} & \includegraphics[width=0.15\linewidth, valign=c]{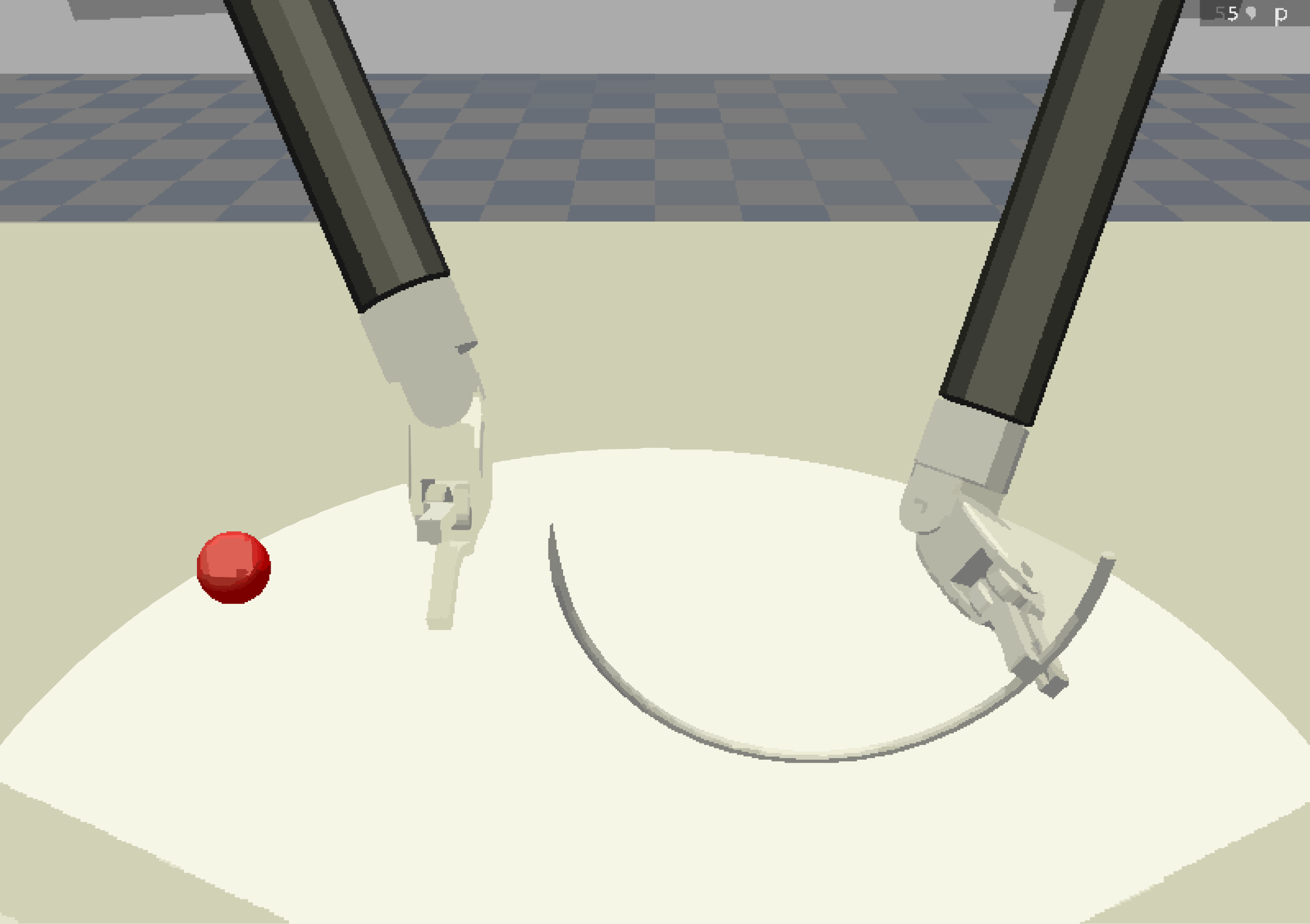} & \includegraphics[width=0.15\linewidth, valign=c]{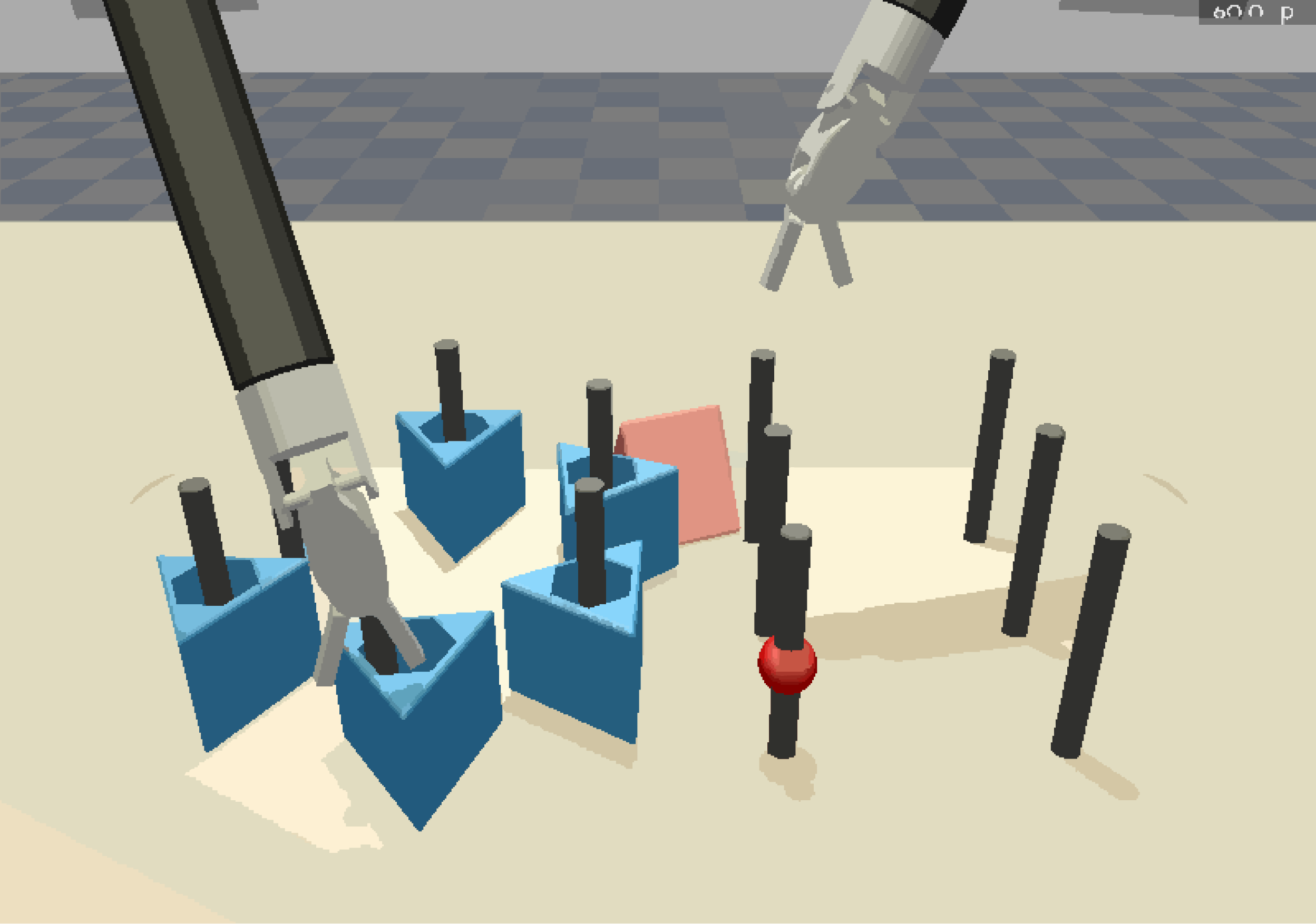} \\
\midrule
BET \cite{shafiullah2022behavior} & $78\%$ & $73\%$ & $\textbf{63}\%$ & --- \\
DEX \cite{huang2023dex}  & $98\%$ & $77\%$ & $56\%$ & --- \\
ALOHA \cite{fu2024mobile} & $95\%$ & $14\%$ & $11\%$ & --- \\
Diffusion \cite{chi2023diffusionpolicy} & $21\%$ & $2\%$ & $2\%$ & --- \\
$\pi_{0.5}$ \cite{black2025pi05} & $8\%$ & $36\%$ & $12\%$ & $33\%$ \\
\midrule
\textbf{Surgical WAM w/o PT (ours)}    & $96\%$ & $66\%$ & $50\%$ & $42\%$ \\
\textbf{Surgical WAM w/ PT (ours)}    & $ \textbf{99}\%$ & $\textbf{86}\%$ & $62\%$ & $\textbf{64}\%$ \\
\bottomrule
\end{tabular}
\vspace{-5mm}
\end{table*}

\paragraph{Stage 2: action-labeled fine-tuning.}
Starting from the video-pretrained initialization $\theta_0$, we
fine-tune the full model on action-labeled surgical demonstrations
$\mathcal{D}_{va}
=
\{(o_{1:T},s_{1:T},a_{1:T},c)\}$. We denote the conditioning context at
time $t$ by $h_t \coloneqq (o_{\leq t},s_{\leq t},c)$. Each training
example pairs this context with an action chunk and the corresponding
future-prediction targets. The action and future-prediction tokens are
inserted into a shared sequence and jointly corrupted according to the
diffusion timestep $\tau$. Let $z_\tau$ denote the resulting noised
sequence. We optimize the standard diffusion denoising objective
\begin{equation}
\mathcal{L}_{\mathrm{ft}}(\theta)
=
\mathbb{E}_{\substack{
(o,s,a,c)\sim\mathcal{D}_{va}\\[-1pt]
\epsilon,\,\tau}}
\left[
\left\|
\epsilon
-
\epsilon_{\theta}(z_{\tau},\tau \mid h_t)
\right\|_2^2
\right].
\label{eq:lft}
\end{equation}
Here, $\epsilon$ denotes the noise added to the shared sequence. Action
and future-prediction tokens are denoised jointly through the same
attention and visual-prediction parameters. This couples the
action-prediction stream to the temporal structure learned during
video-only pretraining, allowing the model to transfer video-derived
dynamics representations while learning to predict executable surgical
actions.

The practical consequence of this recipe is a shift in the data
bottleneck. Stage 1 consumes only action-free video, which is
abundant and inexpensive to collect, while the action-labeled
budget of Stage 2 remains fixed. The recipe thus offers a practical
route to scaling up surgical robot learning: performance grows with
curated surgical video rather than with expensive kinematic data
collection.

\subsection{Closed-Loop Policy Execution}
\label{sec:inference}

At test time the finetuned model is used as a closed-loop
receding-horizon policy. Given the current observation and task
embedding, the diffusion sampler produces an action chunk of length
$H_c$ together with the corresponding future-prediction slots; the
robot executes the first $H_e \le H_c$ actions, discards the rest,
and the loop then re-observes and re-plans. Smaller $H_e$ improves
reactivity while larger $H_e$ amortizes the cost of diffusion
sampling.

\paragraph{Joint sampling of video and action slots.}
Although only actions are sent to the actuator, the model samples the
action and future-prediction tokens jointly: action tokens attend to
the denoised future-prediction tokens at every diffusion step,
preserving at inference the video-action coupling learned during
finetuning. The future-prediction outputs are then discarded, so
joint sampling adds compute but no I/O overhead.

%% file: sec/exp.tex
\section{Experiments}
\label{sec:experiments}

\subsection{Experimental Setup}
\label{sec:exp_setup}

We evaluate the proposed Surgical World--Action Model (\surgwam) on the
\surrol{} surgical manipulation benchmark~\citep{xu2021surrol}.
The benchmark provides endoscopic observations, robot proprioception,
and synchronized actions for simulated \dvrk{} manipulation tasks.

We consider four tasks: Needle Pick, Peg Transfer, Needle Regrasp,
and BiPeg Transfer. Needle Pick and Peg Transfer are unilateral
manipulation tasks, while Needle Regrasp and BiPeg Transfer require
bimanual coordination. Together, these tasks involve reaching,
grasping, contact-rich manipulation, object transfer, and precise
placement.

At each control step, the policy receives the current endoscopic image,
\dvrk{} proprioception, and task embedding. It predicts an action chunk
of length $H_c$, where each action is an absolute Cartesian
end-effector target for the \dvrk{} patient-side manipulators (PSMs),
consisting of a position, an orientation, and a gripper open--close
command per arm. Only the first $H_e$ actions are executed, after which
the robot obtains a new observation and replans. Unless otherwise
specified, we use $H_c=16$ and $H_e=4$.

\paragraph{Action-labeled demonstrations.}
The action-labeled training set contains 10k demonstrations
collected from the \surrol{} simulator \cite{xu2021surrol}. Each demonstration consists of
endoscopic observations, proprioceptive states, task information, and
synchronized robot actions. The same action-labeled dataset is used
for all methods and ablations.

\paragraph{Compared methods.}
We compare the proposed \surgwam against representative
prior policies on the SurRoL benchmark, as well as an ablated
variant of our own model.

\begin{itemize}
    \item \textbf{BET}~\citep{shafiullah2022behavior}: Behavior Transformer, a transformer-based behavior-cloning policy that
    models multi-modal demonstrations by discretizing actions into
    behavior modes.

    \item \textbf{DEX}~\citep{huang2023dex}: a demonstration-guided reinforcement-learning method with efficient exploration, designed for surgical task automation on SurRoL.

    \item \textbf{ALOHA}~\citep{fu2024mobile}: an imitation-learning
    policy with action chunking developed for fine-grained bimanual
    manipulation.

    \item \textbf{Diffusion Policy}~\citep{chi2023diffusionpolicy}: a
    visuomotor policy that generates action sequences via a diffusion model.

    \item $\bm{\pi_{0.5}}$~\citep{black2025pi05}: a large
    vision-language-action model with open-world generalization,
    fine-tuned on the same action-labeled demonstrations.
\end{itemize}

We further compare two variants of our model with identical
architectures and optimization settings:

\begin{itemize}
    \item \textbf{\surgwam w/o PT}: the Cosmos Policy model is
    fine-tuned directly on action-labeled surgical demonstrations
    without additional surgical video pretraining.

    \item \textbf{\surgwam w/ PT}: the model is first pretrained
    on action-free surgical videos and then fine-tuned on the same
    action-labeled demonstrations.
\end{itemize}

The two WAM variants share
the same backbone initialization, action representation, diffusion
sampler, and evaluation protocol; their comparison therefore isolates
the effect of action-free surgical video pretraining.

\paragraph{Evaluation metric.}
We report the closed-loop task success rate. An episode is successful if
the task-specific success condition is satisfied within the maximum
episode horizon. Each result is evaluated over
$100$ evaluation episodes.

\paragraph{Implementation details.}
The model is fine-tuned for $80k$ optimization steps with a batch
size of $1$ and a learning rate of $1e-4$. The diffusion
sampler uses $5$ denoising steps. All
experiments use the same image resolution, optimizer, learning-rate
schedule, and action representation.

\begin{figure*}[!t]
\centering
\includegraphics[width=\textwidth]{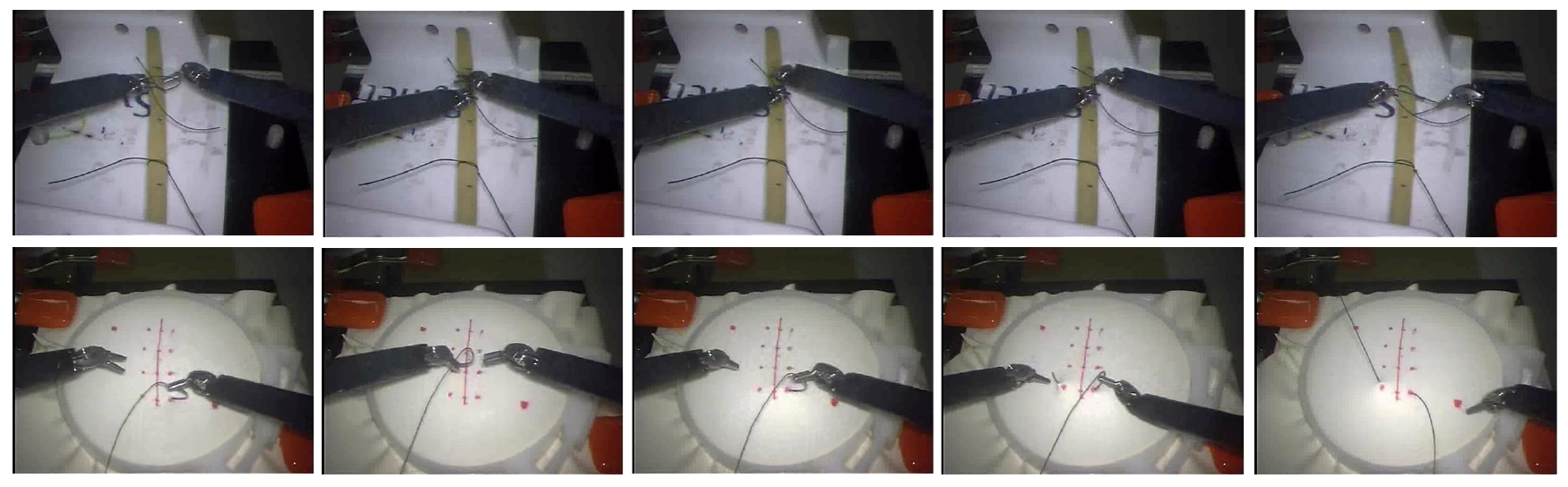}
\vspace{-6mm}
\caption{\textbf{Real-robot surgical manipulation sequences from
JIGSAWS.} Representative endoscopic frames from the JIGSAWS \dvrk{}
teleoperation recordings used in our real-data experiments, with
task progress shown from left to right. \figtop{} Knot tying.
\figbottom{} Suturing.}
\label{fig:jigsaws}
\vspace{-5mm}
\end{figure*}

\subsection{Main Results}
\label{sec:main_results}

\Cref{tab:main} compares the proposed \surgwam against prior
policies on the SurRoL benchmark, together with the corresponding
WAM trained without surgical video pretraining.

\surgwam w/ PT achieves the best success rate on three of the four tasks and remains competitive on the fourth. Notably, the generalist policies transfer poorly to the surgical domain: ALOHA drops to $14\%$ on PegTransfer and Diffusion Policy to $2\%$. Even $\pi_{0.5}$, a state-of-the-art vision-language-action (VLA) model with large-scale multi-domain pretraining, obtains only $8\%$ on NeedlePick and $12\%$ on NeedleRegrasp, averaging $22.3\%$ across the four tasks. While its generalist pretraining yields a moderate $33\%$ on the bimanual BiPegTransfer task, its uniformly low performance on precision-critical tasks suggests that language-conditioned action decoding alone does not capture the fine-grained visual dynamics required in the surgical domain. In contrast, \surgwam grounds action prediction in a learned model of surgical visual dynamics and achieves $64\%$ on BiPegTransfer, nearly doubling the VLA baseline without any language supervision.

Video pretraining further strengthens this advantage. Under the
same action-labeled budget, pretraining improves closed-loop success
on PegTransfer from $66\%$ to $86\%$, an absolute gain of $20$
percentage points. The improvement is consistent across all four
tasks and is most pronounced on the contact-rich and bimanual
problems. Averaged over the full suite, \surgwam w/ PT reaches
$77.8\%$ versus $63.5\%$ without pretraining, an improvement of
$14.3$ points.

These results demonstrate that action-free video pretraining
provides visual-dynamics representations that transfer to action
prediction, even when the downstream action-labeled data are kept
fixed, and that the World-Action model can outperform generalist
VLA pretraining under matched fine-tuning budgets.

\subsection{Ablation Studies}
\label{sec:ablations}

We perform controlled ablations to understand how the benefit of
action-free video pretraining depends on the training budget and the
amount of open-loop execution.
\vspace{-2mm}
\begin{table}[!t]
\caption{\textbf{Fine-tuning-step ablation.} Closed-loop success rate
under different numbers of action-labeled fine-tuning steps.}
\label{tab:ablation_steps}
\centering
\small
\begin{tabular}{l|cc}
\toprule
\textbf{Fine-tuning steps} &
\textbf{WAM w/o PT} &
\textbf{WAM w/ PT} \\
\midrule
$40$k  & $41.5\%$ & $52\%$ \\
$80$k  & $65.5\%$ & $86\%$ \\
$120$k  & $71.5\%$ & $34\%$ \\
$160$k  & $79\%$ & $56.5\%$ \\
\midrule
Best       & $79\%$ & $86\%$ \\
\bottomrule
\end{tabular}
\vspace{-0.2in}
\end{table}

\subsubsection{Effect of Fine-Tuning Steps}
\label{sec:ablation_steps}

We vary the number of Stage 2 optimization steps while keeping the
action-labeled dataset and video-pretraining corpus fixed.

As shown in \Cref{tab:ablation_steps}, the video-pretrained model
peaks at $86\%$ after only $80$k steps, whereas the model without
pretraining requires $160$k steps to reach its best of $79\%$. The
pretrained initialization thus attains a higher peak with half the
optimization budget. At longer horizons, the pretrained model degrades,
which we attribute to overfitting to the limited demonstrations that
gradually erodes the visual-dynamics prior. We therefore report both
the best success rate and the rate at fixed steps, separating faster
convergence from favorable checkpoint selection.
\subsubsection{Effect of the Execution Horizon}
\label{sec:ablation_horizon}

We study the execution horizon $H_e$, the number of predicted actions
executed before replanning. The model always predicts a chunk of fixed
length $H_c = 16$, and we evaluate $H_e \in \{1, 2, 4, 8\}$.

A smaller $H_e$ provides more frequent feedback at the cost of more
diffusion sampling, while a larger $H_e$ amortizes computation but
makes the controller more open-loop. \Cref{tab:ablation_horizon} shows
that the pretrained model is robust across horizons, maintaining at
least $70\%$ success, whereas the model without pretraining is more
sensitive and peaks only at $H_e = 4$ ($66\%$). We adopt $H_e = 4$ as
the default, balancing reactivity against inference cost.

\subsection{Results on Real Surgical Video}
\label{sec:exp_real}

To probe whether the benefit of action-free video pretraining transfers
beyond simulation, we apply the identical two-stage protocol to
demonstrations from the JIGSAWS dataset~\citep{gao2014jigsaws}, a
standard corpus of \dvrk{} operation recordings captured on real robotic hardware
(\Cref{fig:jigsaws}). We observe the same qualitative trend as in
simulation: video-pretrained initializations yield stronger downstream
policies than a from-scratch baseline. This indicates
that the advantage of action-free video pretraining is not an artifact
of the simulator's rendering or dynamics, but persists under the visual
complexity and manipulation difficulty of real surgical scenes.

\vspace{-2mm}
\begin{table}[!t]
\caption{\textbf{Execution-horizon ablation.} Effect of the number of
actions executed before replanning. The predicted chunk length $H_c$
and all model parameters are fixed.}
\vspace{-0.1in}
\label{tab:ablation_horizon}
\centering
\small
\begin{tabular}{lcc}
\toprule
\textbf{$H_e$} &
\textbf{WAM w/o PT} &
\textbf{WAM w/ PT} \\
\midrule
$1$     & $50\%$ & $86\%$  \\
$2$     & $52\%$ & $70\%$ \\
$4$     & $66\%$ & $86\%$  \\
$8$     & $60\%$ & $80\%$ \\
\bottomrule
\end{tabular}
\vspace{-0.2in}
\end{table}

%% file: sec/con.tex
\section{Conclusion}
\label{sec:conclusion}

We presented a two-stage training recipe for surgical manipulation
that decouples visual-dynamics learning from action supervision. In
the first stage, a world action model is pretrained on action-free
surgical videos; in the second stage, the model is fine-tuned on a
fixed set of action-labeled demonstrations. Under an identical
fine-tuning budget, video pretraining improves average closed-loop
success from $63.5\%$ to $77.8\%$ over four simulated tasks, with the
largest gains on contact-rich and bimanual problems. Our ablations
further show that the pretrained initialization converges faster,
reaching a higher peak with half the optimization steps, and remains
robust across execution horizons. Results on real surgical videos
from JIGSAWS confirm that these benefits are not an artifact of
simulation. These findings suggest that abundant unlabeled surgical
video is an effective substitute for costly action labels, and that
scaling action-free pretraining is a promising direction for
data-efficient surgical robot learning.